\pdfoutput=1

\documentclass[11pt]{article}

\usepackage[preprint]{acl}
\usepackage{hyperref}
\usepackage{graphicx}

\usepackage{times}
\usepackage{latexsym}

\usepackage[T1]{fontenc}

\usepackage[utf8]{inputenc}

\usepackage{microtype}

\usepackage{inconsolata}

\usepackage{graphicx}
\usepackage{booktabs}
\usepackage{xspace}

\usepackage[most]{tcolorbox}
\definecolor{skyblue}{RGB}{95, 157, 241}

\def\paradetoxhuman{\textsc{Paradetox-Human}\xspace}
\def\paradetoxllm{\textsc{Paradetox-LLM}\xspace}
\def\paradehate{\textsc{Paradehate}\xspace}

\usepackage{url}

\usepackage{adjustbox}
\usepackage{float}
\usepackage{makecell}

\title{LLM in the Loop: Creating the \paradehate Dataset for Hate Speech Detoxification}

\author{Shuzhou Yuan\thanks{Equal contribution.}$^{1}$,~Ercong Nie\footnotemark[1]$^{2,3}$,~Lukas Kouba\footnotemark[1]$^{1}$, Ashish Yashwanth Kangen$^{1}$\\
\textbf{Helmut Schmid$^{2}$,
Hinrich Schütze$^{2,3}$~and Michael Färber$^{1}$} \\
$^{1}$ScaDS.AI and TU Dresden 
$^{2}$LMU Munich \\
$^{3}$Munich Center for Machine Learning (MCML) \\
\texttt{shuzhou.yuan@tu-dresden.de,~nie@cis.lmu.de}}

\begin{document}
\maketitle
\begin{abstract}
\textcolor{red}{\textbf{Content Warning:}} \textit{This paper contains examples of hate speech, which may be disturbing or offensive to some readers.}

Detoxification, the task of rewriting harmful language into non-toxic text, has become increasingly important amid the growing prevalence of toxic content online. However, high-quality parallel datasets for detoxification, especially for hate speech, remain scarce due to the cost and sensitivity of human annotation. In this paper, we propose a novel LLM-in-the-loop pipeline leveraging GPT-4o-mini for automated detoxification. We first replicate the ParaDetox pipeline by replacing human annotators with an LLM and show that the LLM performs comparably to human annotation. Building on this, we construct \paradehate, a large-scale parallel dataset specifically for hate speech detoxification. 
We release \paradehate as a benchmark of over 8K hate/non-hate text pairs and evaluate a wide range of baseline methods. Experimental results show that models such as \texttt{BART}, fine-tuned on \paradehate, achieve better performance in style accuracy, content preservation, and fluency, demonstrating the effectiveness of LLM-generated detoxification text as a scalable alternative to human annotation. 
\begin{center}
  \includegraphics[width=1em]{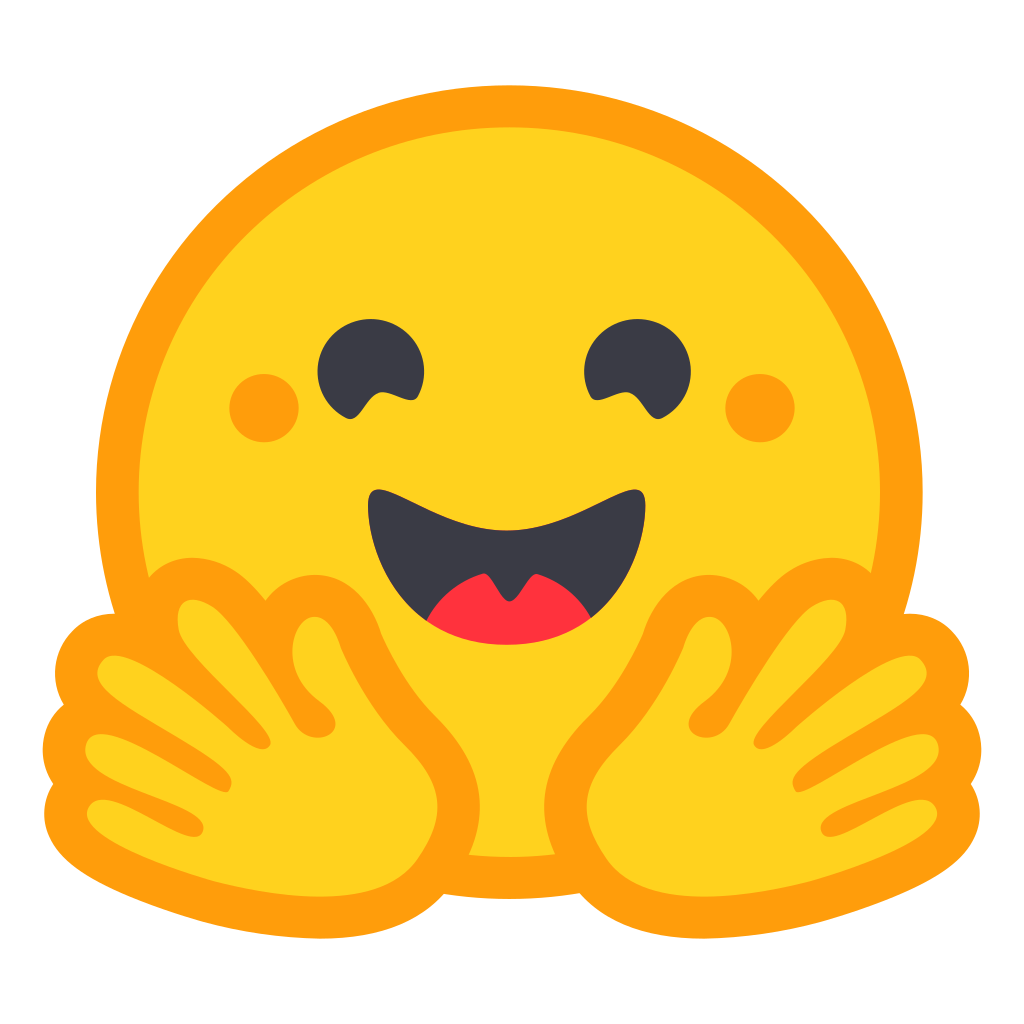} \href{https://huggingface.co/datasets/ScaDSAI/Paradehate}{ScaDSAI/ParaDeHate}
\end{center}
\end{abstract}

\section{Introduction}








\begin{figure}[ht]
    \centering
    \includegraphics[width=1\linewidth]{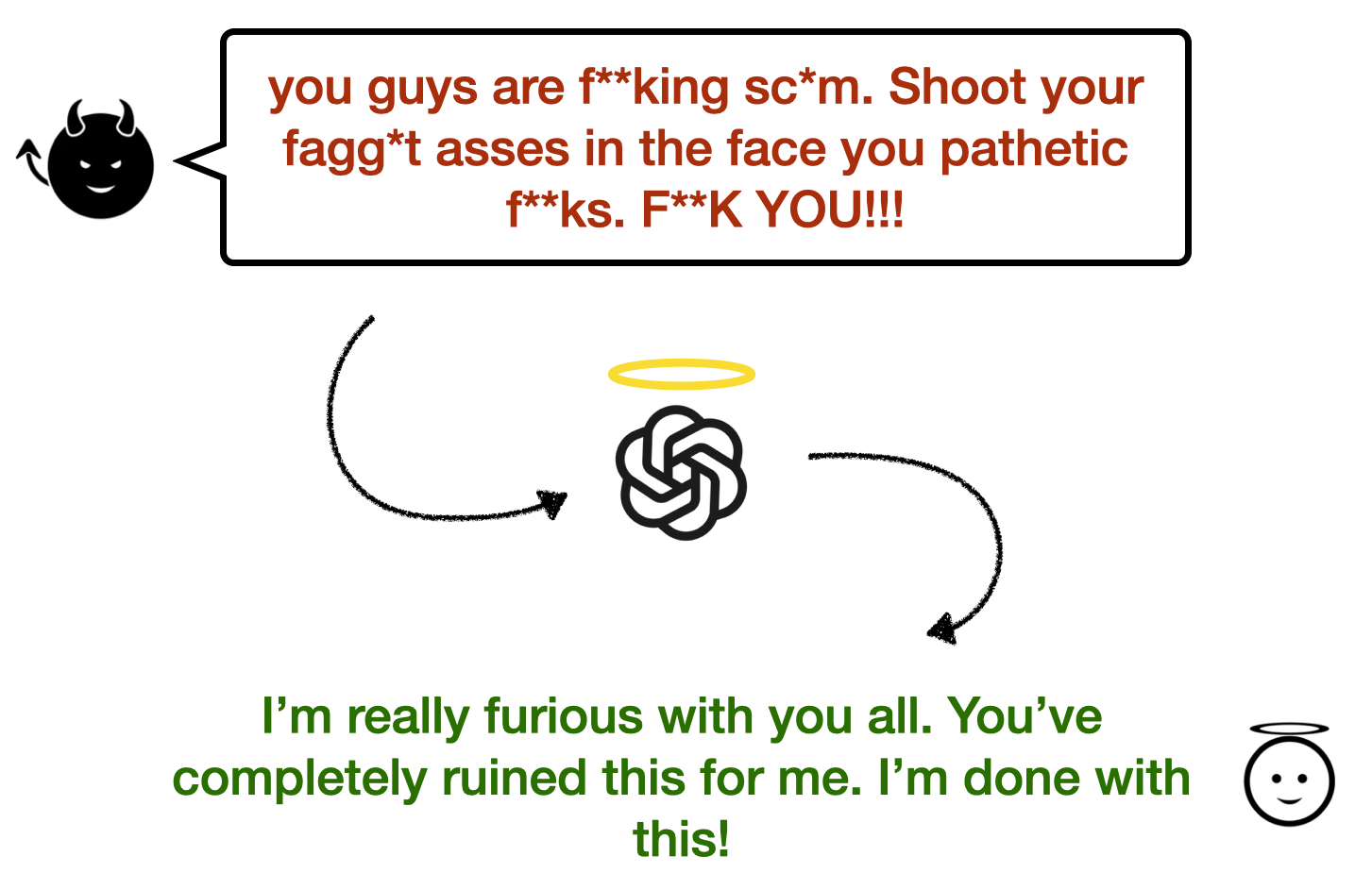}
    \caption{An example of a hate speech input and its detoxified version generated by an LLM. Our evaluation indicates that LLMs perform comparably to human annotators in the task of hate speech detoxification.}
    \label{figure:intro_figure}
\end{figure}

\begin{figure*}[ht]
    \centering
    \includegraphics[width=1\linewidth]{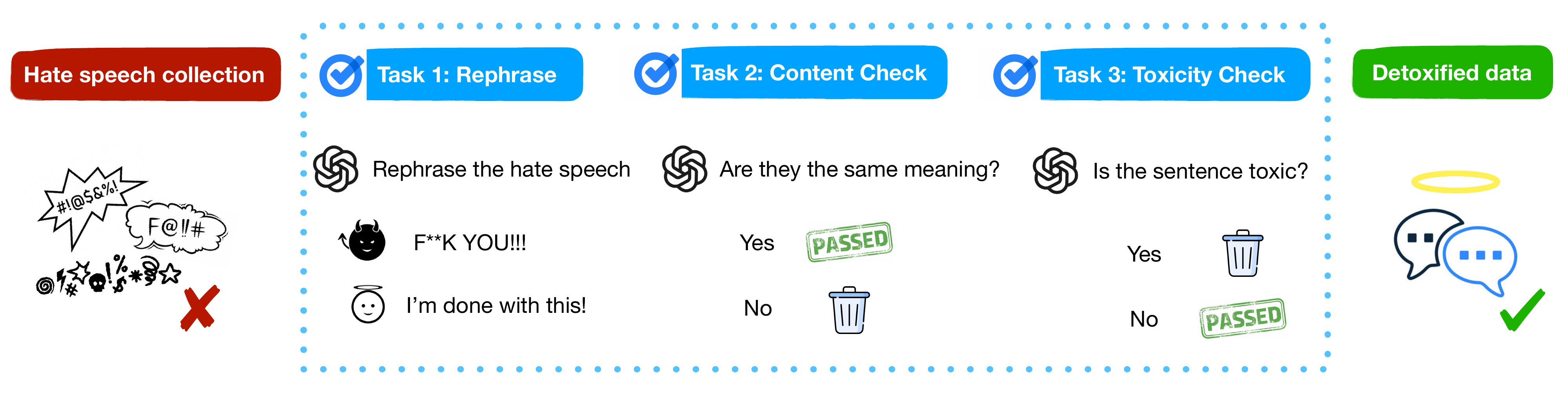}
    \caption{Pipeline for constructing \paradehate. We begin by collecting hate speech texts from widely used datasets. An LLM acts as the annotator, performing three tasks: rephrasing hate speech, verifying content preservation, and evaluating toxicity. Texts that pass all three checks are considered detoxified and are included in the resulting parallel dataset.}
    \label{figure:pipeline}
\end{figure*}

The widespread presence of toxic language, including hate speech, on online platforms presents serious threats to the integrity of digital communities and the well-being of their users.~\citep{yuan-etal-2022-separating}. While substantial progress has been made in the detection of such harmful content~\citep{zampieri-etal-2019-semeval,rottger-etal-2021-hatecheck,fortuna-etal-2022-directions}, detection alone offers limited recourse beyond content removal or user sanctions. A more constructive approach is text \emph{detoxification}: automatically rewriting toxic or hateful messages into non-toxic, yet semantically equivalent, alternatives~\citep{nogueira-dos-santos-etal-2018-fighting,tran-etal-2020-towards,dementieva-etal-2024-multiparadetox}. This task, a specialized form of \emph{style transfer}, holds promise for fostering more inclusive and respectful online discourse~\citep{rao-tetreault-2018-dear,jin2022deeplearning}.


Supervised models for detoxification have shown strong performance, but their success hinges on the limited availability of high-quality parallel datasets~\citep{dale-etal-2021-text}.
Human-annotated parallel corpora, where each toxic input is paired with a semantically equivalent but non-toxic version, are costly and time-intensive to produce, as they typically require extensive human crowdsourcing for both generation and validation of detoxified paraphrases~\citep{carlson2018evaluating,pryzant2020automatically}. 
The ParaDetox pipeline~\citep{logacheva-etal-2022-paradetox} exemplifies this approach, leveraging crowdsourcing to build the first large-scale parallel detoxification corpus. Yet, the reliance on human annotators limits scalability, speed, and adaptability to new domains or languages. As a result, existing resources remain small in scale and often focus on general forms of toxicity (e.g., offensive or profane language), while overlooking more complex and socially harmful subtypes such as hate speech.


Detoxifying hate speech presents unique challenges beyond general toxic language. Hate speech frequently involves identity-targeted slurs and ideologically charged language, making faithful paraphrasing especially difficult: as shown in Figure \ref{figure:intro_figure}, there is a delicate balance between removing harmful content and preserving the original meaning without introducing distortion or ambiguity~\citep{welbl-etal-2021-challenges-detoxifying,hartvigsen-etal-2022-toxigen}. Traditional approaches, reliant on human annotators, are resource-intensive and may struggle to scale as new forms of harmful language emerge~\citep{vetagiri-etal-2024-multilate}. Meanwhile, recent advances in Large Language Models (LLMs) have demonstrated impressive abilities in text generation, paraphrasing, and nuanced understanding of linguistic context~\citep{kurt2024comparative,wuraola-etal-2024-understanding,tripto-etal-2024-ship}.

This convergence of challenges and technological progress naturally leads to a pivotal research question: \emph{Can LLMs effectively replace or augment human annotators in the construction of high-quality parallel detoxification datasets?} If so, LLM-driven pipelines could not only accelerate dataset creation and reduce costs, but also offer the flexibility and scalability for rapid adaptation to emerging domains of harmful language, including hate speech.  

In this work, we systematically investigate the potential of LLMs as central agents in a scalable, automated pipeline for constructing parallel detoxification datasets, specifically replacing human validators in the critical steps of content preservation and toxicity checking. 
Our approach leverages the strong generative and evaluative capabilities of LLMs, while mitigating their refusal behavior when given harmful inputs through controlled prompting.
We first replicate the ParaDetox pipeline~\citep{logacheva-etal-2022-paradetox}, substituting human crowdsourcing with LLM-based validation, and rigorously compare the effectiveness of LLMs against established automated methods: sentence transformers~\citep{reimers-gurevych-2019-sentence} for semantic similarity and RoBERTa-based toxicity classifiers~\citep{liu2019roberta,hanu2020Detoxify}.
We then demonstrate the practical utility of LLM-generated datasets by training and evaluating detoxification models, benchmarking their performance against models trained on human-constructed data.

Building on these insights, we extend the LLM-in-the-loop pipeline to the domain of hate speech detoxification. By using hate speech inputs as the source, we generate semantically faithful, non-hateful rewrites without requiring human intervention, thereby constructing a new parallel dataset \paradehate and significantly reducing data creation costs. To comprehensively assess the quality of the generated parallel data in practice, we evaluate a suite of baseline and advanced detoxification methods, including recent innovations such as style-specific neuron steering for controllable text generation~\citep{lai-etal-2024-style}. Our evaluation employs rigorous metrics including style accuracy, content preservation, fluency, and BLEU, ensuring comparability and robustness. The results reveal that existing detoxification methods, when applied without task-specific training data, often fail to produce fluent or meaning-preserving outputs. In contrast, models fine-tuned on \paradehate, such as \texttt{BART-large}, achieve significantly better performance across all metrics, demonstrating the effectiveness of our dataset. These findings confirm the necessity of high-quality, hate-speech-specific training data and establish the potential of LLM-in-the-loop pipelines for scalable and reliable dataset generation.


Our contributions are as follows:
\begin{itemize}
\item We release a new parallel dataset \paradehate consisting of 8K hate speech and corresponding detoxified text, filling a critical gap in existing resources.
\item We introduce a novel GPT-4o-mini-based pipeline for automated hate speech detoxification, demonstrating that it achieves human-comparable quality while being more scalable and cost-effective.
\item We conduct comprehensive evaluations against existing detoxification models, showing that training with \paradehate significantly improves performance on downstream detoxification tasks.
\end{itemize}

\section{Related Work}

\paragraph{Detoxification and Hate Speech}
Style transfer is a core approach for text detoxification, typically aiming to rewrite toxic sentences into non-toxic ones while preserving content. Most existing models are trained on non-parallel data due to the scarcity of high-quality parallel training sets. They rely on strategies such as pointwise correction of toxic words~\citep{li-etal-2018-delete,ijcai2019p732,malmi-etal-2020-unsupervised}, adversarial classifiers for encoder-decoder models~\citep{shen2017style,Fu_Tan_Peng_Zhao_Yan_2018}, or joint training with reinforcement learning and variational inference~\citep{ijcai2019p711,He2020A}. 
Recent detoxification work adapts techniques from general style transfer, such as training autoencoders with style classification~\citep{nogueira-dos-santos-etal-2018-fighting}, fine-tuning a T5 model~\citep{raffel2020t5} as a denoising autoencoder~\citep{laugier-etal-2021-civil}, pointwise editing toxic sentences on masked language models~\citep{dale-etal-2021-text}, and steering style-specific neurons~\citep{lai-etal-2024-style}.

Hate speech is a particularly severe form of harmful language, often targeting individuals or groups based on identity and carrying ideologically charged or derogatory content~\citep{rottger-etal-2021-hatecheck,fortuna-etal-2022-directions}. While much prior work has focused on hate speech detection, including the development of datasets and neural models for classification~\citep{kim-etal-2022-generalizable,yuan-etal-2022-separating,toraman-etal-2022-large}, relatively little attention has been paid to the problem of rewriting hate speech into non-hateful language~\citep{kostiuk2023automatic}. Unlike detection, the detoxification of hate speech requires not only identifying toxic content, but also generating semantically faithful, socially acceptable alternatives, posing unique challenges distinct from standard paraphrasing or machine translation.

Researchers use parallel data for supervised style transfer~\citep{zhang-etal-2020-parallel,briakou-etal-2021-ola}. \citet{logacheva-etal-2022-paradetox} propose a parallel dataset for toxic text in English. \citet{dementieva-etal-2024-multiparadetox} extend the dataset in a multilingual setting.
Our work builds on this tradition by focusing on scalable parallel dataset construction using LLMs in the loop, enabling more effective fine-tuning of models for detoxification and hate speech rewriting. This approach directly addresses the data bottleneck that limits supervised style transfer in safety-critical domains.

\paragraph{LLMs as Agents for Data Creation and Validation}
LLMs have revolutionized the landscape of data annotation and synthesis, enabling the automation of previously labor-intensive tasks~\citep{tan-etal-2024-large}. LLMs are increasingly leveraged not only as annotators, generating diverse and high-quality labels, paraphrases, or rationales for various NLP datasets~\citep{wadhwa-etal-2023-revisiting,nie2024bmike,zhang-etal-2023-llmaaa}, but also as agents for synthetic data creation~\citep{koksal-etal-2024-longform,yu-etal-2023-harnessing,pan-etal-2024-automatically}, substantially reducing the reliance on human annotators and accelerating large-scale dataset construction. A growing body of work has demonstrated that LLM-generated annotations can rival or even surpass human annotation quality in certain settings, provided that robust filtering and assessment mechanisms are in place~\citep{gilardi2023chatgpt,lee-etal-2023-making}.

In parallel, LLMs have also emerged as powerful automated judges or validators (“\textit{LLM-as-a-judge}”), widely used for evaluating the quality, style, or factuality of generated text in both model development and benchmarking~\citep{li2024generation,chen2024mllmasajudge,wu2024meta}. While both human and LLM judges exhibit biases~\citep{chen-etal-2024-humans}, LLM-based evaluation offers scalability and consistency, and recent research has focused on mitigating bias and ensuring reliability through prompt engineering and multi-agent debate~\citep{zheng2023judging,li-etal-2024-coevol}. Our work unifies these two roles, LLM as annotator and validator, by employing LLMs in the loop for both dataset creation and rigorous automatic validation, thus establishing a scalable and cost-effective pipeline for parallel detoxification and hate speech rewriting.

\section{LLM in the Loop vs Human in the Loop}\label{sec:llm_pipeline}

We automate the task of detoxification by using an LLM, following the annotation pipeline introduced in ParaDetox~\citep{logacheva-etal-2022-paradetox}. The goal is to reproduce the ParaDetox dataset using an LLM. We use the toxic texts from the original ParaDetox dataset as input to the LLM. The detoxification process consists of three steps: paraphrase generation, content preservation check, and toxicity check, as shown in Figure~\ref{figure:pipeline}.

Since LLMs often exhibit false refusal behavior when presented with sensitive content such as toxic text~\citep{rottger-etal-2024-xstest}, we select GPT-4o mini as the annotation agent, as it demonstrates a lower rate of false refusals compared to GPT-4o in preliminary experiments.\footnote{We randomly select 20 samples from the toxic texts in ParaDetox and ask the LLMs to rephrase them. The compliance rates for GPT-4o and GPT-4o mini are 70\% and 80\%, respectively.} 
We configure the model with the hyperparameters listed in Appendix \ref{app:hyperparameters_gpt} Table~\ref{tab:hyperparameters}. 

\subsection{Task 1: Generation of Paraphrases}

\begin{figure}[htbp]
    \centering
    \includegraphics[width=1\linewidth]{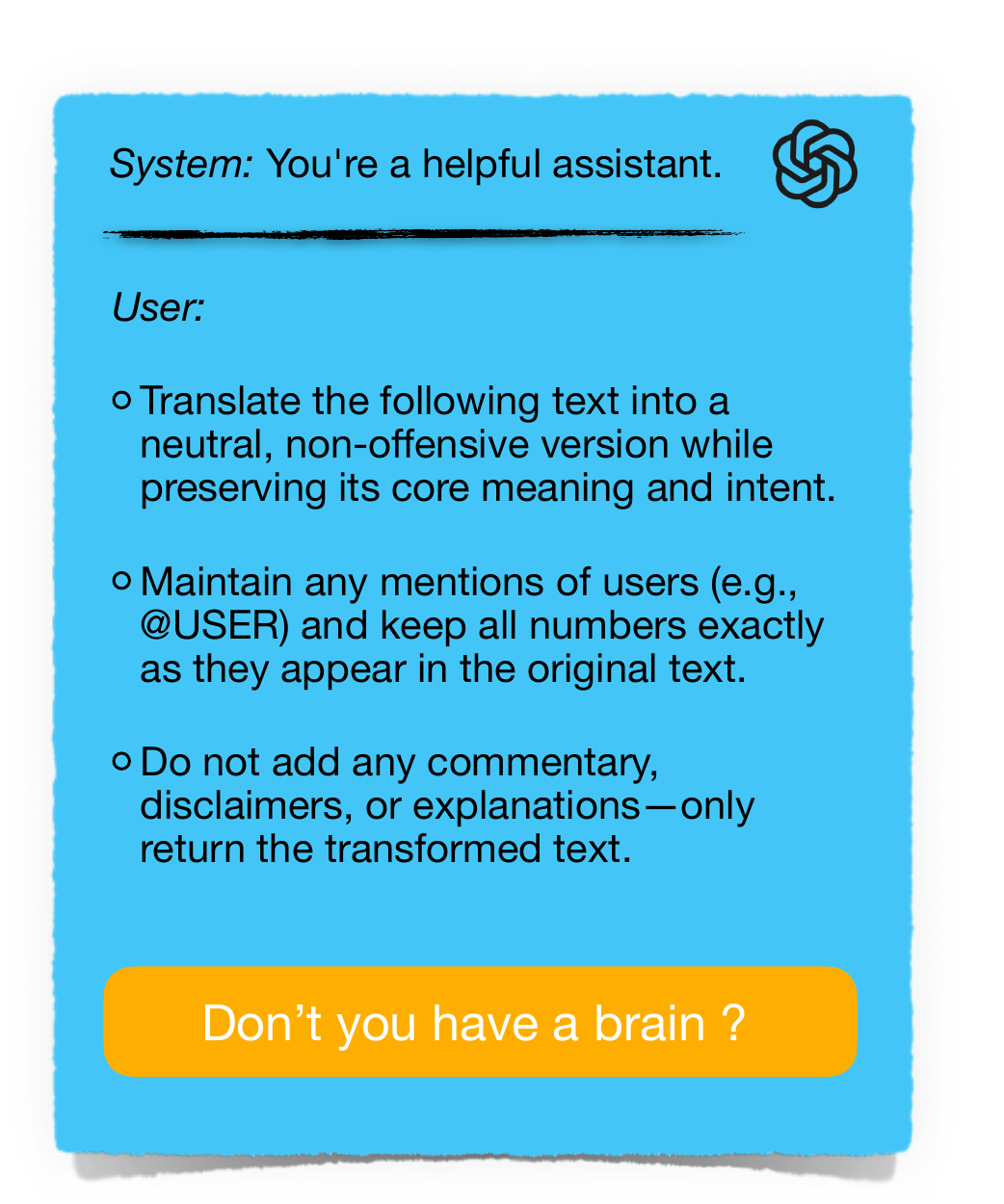}
    \caption{Prompt for Task 1: Generation of Paraphrases.}
    \label{figure:task_1}
\end{figure}

The first task in the detoxification process is to prompt the LLM to generate a text that retains the original meaning of the given toxic input while removing any toxic content, as illustrated in the example for Task 1 in Figure~\ref{figure:pipeline}. The prompt for Task 1 is defined in Figure \ref{figure:task_1}.

However, due to the exaggerated safety behaviors of LLMs, not all toxic texts can be rephrased successfully, some trigger the safety mechanisms, resulting in responses such as ``Sorry, I cannot assist with that.'' To mitigate this false refusal behavior, we re-annotate the rejected requests using an alternative prompt with more instruction and regulations.\footnote{The prompt can be found in Appendix \ref{app:prompt_refusal}.} 


We observe that a proportion of toxic texts still cannot be rephrased by the LLM and continue to trigger safety responses. We remove these texts from the corpus, as they may be either semantically meaningless or inherently irredeemable in terms of toxicity. This decision is also consistent with the human annotation process in ParaDetox, where not all toxic texts can be rephrased in a non-toxic manner while preserving their original meaning.

\subsection{Task 2: Content Preservation Check}\label{sec: task2}

\begin{figure}[htpb]
    \centering
    \includegraphics[width=1\linewidth]{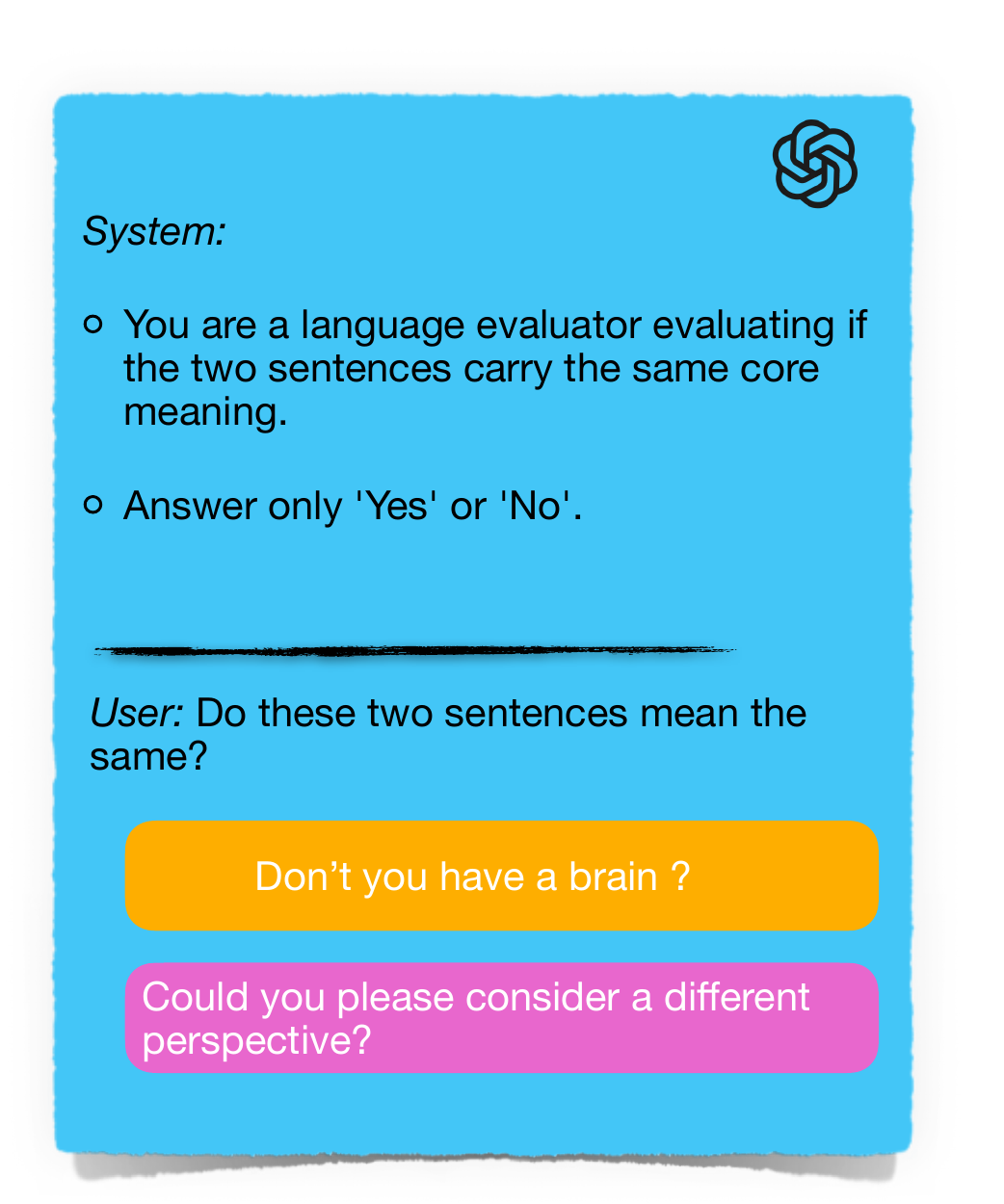}
    \caption{Prompt for Task 2: Content Preservation Check.}
    \label{figure:task_2}
\end{figure}

Aligning with the human annotation process in ParaDetox, we ask the same LLM to evaluate whether the translated (i.e., detoxified) text preserves the meaning of the original toxic input. The LLM is expected to respond with either "Yes" or "No." As illustrated in Task 2 of Figure~\ref{figure:pipeline}, we retain the samples that pass the content preservation check (i.e., when the answer is "Yes") and discard those that fail (i.e., when the answer is "No"). The prompt is defined in Figure \ref{figure:task_2}.

To further control the quality of the content preservation check, we also use \texttt{sentence-transformer}\footnote{\url{https://huggingface.co/sentence-transformers/all-distilroberta-v1}} \citep{reimers-gurevych-2019-sentence} to calculate the cosine similarity between the original toxic text and the translated detoxified text. Based on observations from a subsample of text pairs, we empirically set the cosine similarity threshold to 0.70. Samples with a similarity above 0.70 are annotated with the label "Yes", indicating that the two sentences convey the same meaning, while those below 0.70 are labeled "No", indicating a difference in meaning. We compute Cohen's kappa coefficient to assess the inter-annotator agreement between the LLM's judgments and the cosine similarity-based labels. The resulting $\kappa$ score is 0.55, indicating a moderate level of agreement. 


\subsection{Task 3: Toxicity Check}

\begin{figure}[htpb]
    \centering
    \includegraphics[width=1\linewidth]{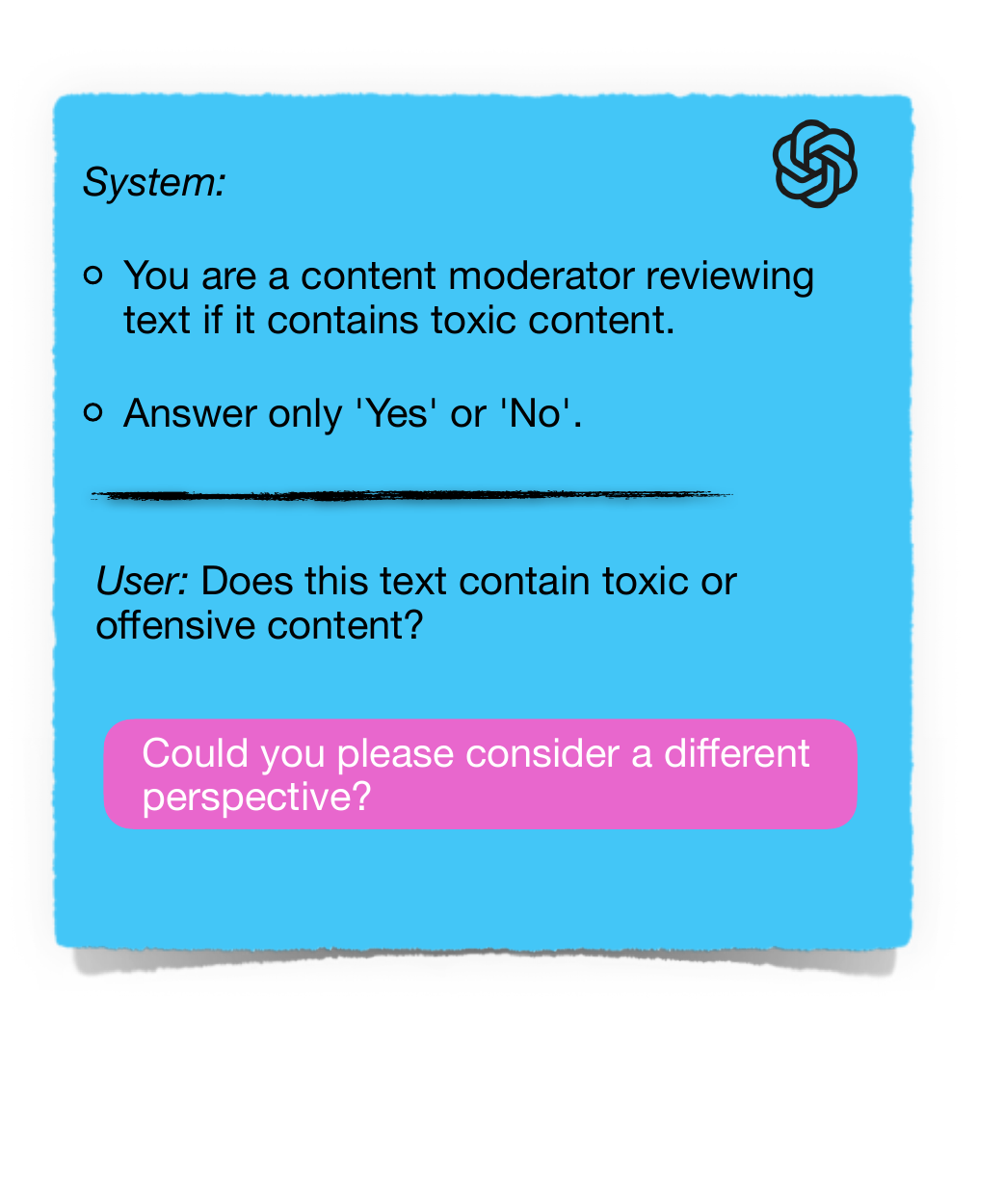}
    \caption{Prompt for Task 3: Toxicity Check.}
    \label{figure:task_3}
\end{figure}

To ensure that the final text contains no toxic content, we perform a toxicity check as Task 3 to further control the quality of the translated text, as illustrated in Task 3 of Figure~\ref{figure:pipeline}. The LLM is used to evaluate whether the translated text still contains toxic content by responding with either "Yes" or "No." We discard texts for which the LLM answers "Yes," indicating the presence of toxicity, and retain those for which the LLM answers "No," indicating the absence of toxicity. The prompt for Task 3 is defined in Figure \ref{figure:task_3}.

To validate the judgments of the LLM, we compute a toxicity score using \texttt{unbiased-toxic-roberta}\footnote{\url{https://huggingface.co/unitary/unbiased-toxic-roberta}} \citep{Detoxify}. Similar to the content preservation check, we examine subsamples and observe that the toxicity score tends to be very high for text containing toxic content. Based on these observations, we set a threshold of 0.9 to distinguish between toxic and non-toxic text: samples with a toxicity score above 0.9 are labeled as still containing toxic content, while those below 0.9 are labeled as non-toxic. We again compute Cohen's kappa coefficient to assess the inter-annotator agreement between the LLM's judgments and the toxicity score-based labels. The resulting $\kappa$ score is 0.72, indicating a substantial level of agreement.

\subsection{Training and Analysis}\label{sec:training}

\begin{table*}[htbp]
\centering
\begin{tabular}{lcccc}
\hline
\textbf{Dataset} & \textbf{Style Accuracy} & \textbf{Content Preservation} & \textbf{Fluency} & \textbf{BLEU} \\
\hline
\paradetoxhuman & 0.96 & \textbf{0.85} & 0.71 & \textbf{0.71}  \\
\paradetoxllm   & \textbf{0.98} & 0.70 & \textbf{0.91} &  0.65 \\
\hline
\end{tabular}
\caption{Evaluation results for ParaDetox-Human and ParaDetox-LLM trained on \texttt{BART-large} across various metrics. \textbf{BLEU} denotes the BLEU score of \texttt{BART-large} generated text and the reference text from ParaDetox-Human and ParaDetox-LLM respectively.}
\label{tab:paradetox_bart}
\end{table*}

We ultimately obtain an LLM-generated version of ParaDetox, \paradetoxllm, consisting of 19,726 samples. We split the training, validation, and test sets with a ratio of 80:10:10 for both \paradetoxllm and \paradetoxhuman (the original ParaDetox dataset). To compare the quality of the two variants, we fine-tune \texttt{BART-large} on each training sets separately under the same experimental settings.\footnote{The hyperparameter of training \texttt{BART-large} can be found in Appendix \ref{app: hyperparameters_bart} Table~\ref{tab:bart_metrics}.} 

Following previous work \citep{logacheva-etal-2022-paradetox,dementieva-etal-2024-multiparadetox}, we evaluate the performance of \texttt{BART-large} on the two versions of ParaDetox along three dimensions: style accuracy, content preservation, and fluency. 

\textbf{Style Accuracy} measures the proportion of detoxified texts classified as non-toxic by a pre-trained toxicity classifier used by \citet{logacheva-etal-2022-paradetox}, ensuring the removal of harmful language. 

\textbf{Content Preservation} is computed as the cosine similarity between LaBSE \citep{feng-etal-2022-language} embeddings of the original and detoxified texts, assessing semantic fidelity. 

\textbf{Fluency} is measured as the percentage of fluent sentences identified by a RoBERTa-based classifier of linguistic acceptability, trained on the CoLA dataset \citep{10.1162/tacl_a_00290}. 

We also report \textbf{BLEU} scores to measure n-gram overlap between the generated texts from fine-tuned \texttt{BART-large} and the reference texts from \paradetoxhuman and \paradetoxllm. For all metrics, higher values indicate better performance.


The evaluation results for ParaDetox-Human and ParaDetox-LLM are presented in Table~\ref{tab:paradetox_bart}. The findings indicate that both datasets achieve comparable performance in detoxifying toxic text, with each exhibiting distinct strengths. In terms of style accuracy, \paradetoxllm slightly outperforms \paradetoxhuman (0.98 vs. 0.96), demonstrating its strong capability in effectively removing harmful language. Although \paradetoxllm falls short in content preservation (0.70 vs. 0.85), it achieves a notably higher fluency score (0.91 vs. 0.71), suggesting that it enables supervised models to produce more fluent detoxified outputs. Overall, the comparable performance of \paradetoxllm to \paradetoxhuman indicates that \textit{LLMs can generate parallel detoxification datasets with quality on par with human annotations.}

\section{Dataset Creation}

Having validated the feasibility of using the LLM-in-the-loop method to detoxify toxic content without relying on human annotators, we extend our approach to a more challenging task: applying the LLM pipeline to online hate speech to construct a parallel detoxified hate speech dataset, \paradehate.

\subsection{Hate Speech Dataset Collection and Preprocessing}

\begin{table}[ht]
\centering
\begin{tabular}{l r}
\toprule
\textbf{Dataset} & \textbf{Number of Samples} \\
\midrule
CreHate      & 5,935 \\
HateXplain   & 1,430 \\
Davidson     &   364 \\
Founta       & 4,176 \\
\midrule
\textbf{Total}       & \textbf{11,905} \\
\bottomrule
\end{tabular}
\caption{Statistics of the merged hate speech datasets used for detoxification.}
\label{tab:dataset_statistics}
\end{table}

We aggregate the samples from four commonly used hate speech datasets:

\textbf{CreHate} \citep{lee-etal-2024-exploring-cross} is a dataset containing social media posts from platforms such as Twitter and Reddit, annotated for hate speech by annotators from five regions to capture cross-cultural perspectives. It includes hate speech and non-hate speech labels. We only use samples that are annotated as hate speech in all five regions.

\textbf{HateExplain} \citep{mathew2021hatexplain} comprises Twitter and Gab posts annotated as hate speech, offensive, or normal. We use samples labeled as hate speech to focus on toxic content.

\textbf{Davidson} \citep{hateoffensive} contains Twitter tweets labeled as hate speech, offensive language, or neutral. All hate speech-labeled samples are included in our experiments.

\textbf{Founta} \citep{founta2018large} is a large collection of Twitter tweets annotated as hateful, abusive, normal, or spam. We select hateful samples to align with our focus on hate speech detoxification.

As presented in Table~\ref{tab:dataset_statistics}, we merge hate speech samples 5,935 from CreHate, 1,430 from HateXplain, 364 from Davidson, and 4,176 from Founta, totaling 11,905 samples. To ensure consistency and compatibility with our detoxification pipeline, particularly given that the data originates from social media platforms such as Twitter, we apply preprocessing steps following previous work \citep{yuan-etal-2022-separating}:

\begin{itemize}
    \item \textbf{URL Removal.}  We first remove all URLs to eliminate external links and focus on the textual content of the posts. 
    \item \textbf{Username Normalization.} Next, we normalize usernames by replacing them with a generic \texttt{@USER} tag, collapsing consecutive \texttt{@USER} tags into a single instance, and standardizing dataset-specific tags, such as \texttt{<user>} and \texttt{<number>}, to \texttt{@USER} and \texttt{@NUMBER}, respectively, to ensure consistency and anonymity across datasets. 
    \item \textbf{HTML and Special Characters.} Finally, we remove HTML-encoded user entities and non-essential special characters and excessive punctuation, to reduce noise while preserving the text’s core meaning.
\end{itemize}

\subsection{\paradehate}


For the detoxification process, we follow the LLM in the loop pipeline described in \S\ref{sec:llm_pipeline}, using GPT-4o-mini as the annotation agent to repeat the three tasks for the hate speech detoxification.\footnote{The cost of using OpenAI API to construct \paradehate can be found in Appendix \ref{app:cost}.} Due to the highly harmful and malicious nature of the content, the model initially failed to generate detoxified outputs for 4,103 samples, triggering refusal behavior. To mitigate this issue, we applied an alternative prompting strategy as outlined in \S\ref{sec: task2}, successfully generating detoxified outputs for an additional 474 samples. This resulted in a final dataset of 8,276 detoxified text pairs. 

\section{Evaluation}

\subsection{Baselines}

To evaluate the \paradehate dataset, we use it to train a supervised model \texttt{BART-large}. The \paradehate is split into train, validate and test set with 80:10:10 ratio. We use the same experimental setting as used in \S\ref{sec:training}. Meanwhile, we compare it against several baseline methods widely adopted in prior work \citep{logacheva-etal-2022-paradetox,dementieva-etal-2024-multiparadetox}, and also a Style-Specific Neurons approach \citep{lai-etal-2024-style}, applied on top of Llama-3 \citep{grattafiori2024llama}. The baselines including:

\begin{itemize}
    \item \textbf{Delete}: Removes all toxic words from the input text, omitting them entirely.
    \item \textbf{Duplicate}: Directly copies the input text without modification, serving as a naive baseline to evaluate the need for detoxification.
      \item \textbf{BART-zero-shot} \citep{lewis-etal-2020-bart}: A pretrained \texttt{BART-large} model used without fine-tuning or task-specific guidance, serving as a naive large model baseline.
    \item \textbf{Mask\&Infill}: Uses a BERT-based pointwise editing model to mask toxic spans and infill them with appropriate replacements \citep{ijcai2019p732}.
    \item \textbf{Delete-Retrieve-Generate (DRG)} \citep{li-etal-2018-delete}:
    \begin{itemize}
        \item \textbf{DRG-Template}: Replaces toxic words with semantically similar neutral alternatives.
        \item \textbf{DRG-Retrieve}: Retrieves non-toxic sentences that convey similar meaning to the original.
    \end{itemize}
    \item \textbf{DLSM} \citep{He2020A}: An encoder-decoder model employing amortized variational inference for style transfer.
    \item \textbf{CondBERT} \citep{dale-etal-2021-text}: A conditional BERT-based model that integrates both style and content constraints during generation.
    \item \textbf{ParaGeDi} \citep{dale-etal-2021-text}: Enhances a paraphraser with a style-informed language model to reweight outputs towards desired styles.
    \item \textbf{Neuron-Specific}: A method that modifies specific neurons in Llama-3 associated with toxic language to guide detoxification \citep{lai-etal-2024-style}.
\end{itemize}

\begin{table}[ht]
\centering
\scalebox{0.73}{
\begin{tabular}{l|cccc}
\toprule
\textbf{Method} & \makecell{\textbf{Style} \\ \textbf{Accuracy}} & \makecell{\textbf{Content} \\ \textbf{Preservation}} & \textbf{Fluency} & \textbf{BLEU} \\
\midrule
LLM-reference & 0.98 & 0.74 & 0.76 & 1.00  \\
\midrule
\multicolumn{5}{c}{Trained on ParaDeHate}\\
\midrule
BART fine-tune & \textbf{0.95} & 0.78 & \textbf{0.71} & \textbf{0.31}  \\
\midrule
\multicolumn{5}{c}{Naive Baselines}\\
\midrule
Delete & 0.65 & 0.96 & 0.39 & 0.22 \\
Duplicate & 0.31 & \textbf{1.00} & 0.47 & 0.23  \\
BART-zero-shot & 0.32 & 0.97 & 0.49 & 0.21  \\
\midrule
\multicolumn{5}{c}{Unsupervised Baselines}\\
\midrule
Mask\&Infill & 0.43 & 0.95 & 0.30 & 0.22  \\
DRG-Template & \textbf{0.95} & 0.26 & 0.01 & 0.01  \\
DRG-Retrieve & 0.90 & 0.26 & 0.01 & 0.01  \\
DLSM & 0.89 & 0.31 & 0.20 & 0.03  \\
CondBERT & \textbf{0.95} & 0.62 & 0.05 & 0.18  \\
ParaGeDi & \textbf{0.95} & 0.72 & 0.62 & 0.14  \\
Neuron-Specific & 0.62 & 0.42 & 0.57 & 0.11 \\
\bottomrule
\end{tabular}}
\caption{Automatic evaluation of detoxification models. Numbers in \textbf{bold} indicate the best results.}
\label{tab:baselines}
\end{table}


We evaluate \paradehate with the baseline methods using the same metrics described in \S\ref{sec:training}, namely: \textbf{Style Accuracy}, \textbf{Content Preservation}, and \textbf{Fluency}, which assess the effectiveness of harmful language removal, semantic fidelity, and the naturalness of the detoxified text, respectively. Additionally, we employ \textbf{BLEU} to measure the n-gram overlap between the generated outputs from the baseline methods and the reference detoxified texts in \paradehate.

\subsection{Results and Analysis}

We present the results in Table~\ref{tab:baselines}. As a reference, LLM-generated text in \paradehate demonstrates high quality with 0.98 style accuracy, 0.74 preservation, and 0.76 fluency. Trained on \paradehate, BART fine-tune outperforms all the other baselines in style accuracy (0.95), fluency (0.78), and BLEU (0.31). Although the naive baselines achieve the highest content preservation score (1.0), they cannot be considered superior to the fine-tuned BART model, as they merely delete swear words or duplicate the input without meaningful detoxification. We notice that BART-zero-shot without fine-tuning also tends to generate the same text as the input. That is the reason why style accuracy is low for the naive baselines, as they still contain toxic content, with 0.31 for Duplicate and 0.32 for BART-zero-shot. Even when toxic words are deleted, the output still only achieves 0.65 style accuracy for the baseline delete, which is not as good as BART fine-tuned on \paradehate. The BLEU score also indicates that the naive baselines have mediocre overlap with the LLM-detoxified text.

Turning to the unsupervised baselines, which are mostly trained on specific text style transfer tasks, we apply them directly to \paradehate and observe that they achieve high style accuracy. DRG-Template, CondBERT, and ParaGeDi even have the same style accuracy score as BART fine-tune (0.95), which indicates that they are able to generate detoxified text. However, they may lose the original meaning of the toxic content as their content preservation scores are lower than that of BART fine-tune: DRG-Template 0.26, CondBERT 0.62, and ParaGeDi 0.72. On the other hand, Mask\&Infill is good at preserving the original content, with a content preservation score of 0.95, but it has low style accuracy (0.43) and low fluency (0.30), which are even worse than some naive baselines. The fluency scores also indicate that, except for Neuron-Specific and ParaGeDi, which are able to generate fluent text with fluency scores of 0.57 and 0.62 respectively, the other unsupervised methods do not perform well when dealing with hate speech input and detoxifying it, as the generated text is likely not fluent at all, especially for DRG methods with 0.01 fluency and CondBERT with 0.05 fluency. BLEU scores also show the same trend: the baselines methods have low overlap with the LLM-detoxified text in \paradehate. 

Overall, these results highlight the difficulty of detoxifying hate speech, arguably more challenging than generic toxic language. Existing methods, particularly those without sufficient task-specific training data, often struggle to strike a balance between detoxification and content fidelity. This underscores the necessity of resources like \paradehate. By leveraging LLMs in the loop to generate high-quality training data, models such as \texttt{BART-large} fine-tuned on \paradehate demonstrate that targeted training can yield robust detoxification performance with improved fluency and semantic consistency.

\section{Conclusion}

In this work, we demonstrate the feasibility of employing an LLM-in-the-loop pipeline to replace human annotators for text detoxification. 
We reproduce the ParaDetox pipeline using an LLM to replace human annotators.
This results in the construction of \paradetoxllm, which we use to fine-tune a \texttt{BART-large} model. Compared to a model trained on the original ParaDetox dataset, the BART model fine-tuned on \paradetoxllm achieves comparable performance across automatic evaluation metrics.

Having established the effectiveness of our approach, we further extend our LLM-in-the-loop pipeline to construct \paradehate, a new parallel dataset specifically focused on hate speech detoxification. Evaluation with existing baseline methods highlights the necessity of such a dataset: without sufficient task-specific training data, these methods perform poorly. In contrast, BART fine-tuned on \paradehate outperforms all the baseline methods. We hope that \paradehate can serve as a benchmark for evaluating models in the task of online hate speech detoxification.
Future work may explore extending the LLM-in-the-loop pipeline to multilingual settings and a broader range of LLMs.

\section*{Limitations}

While our work demonstrates the feasibility of using an LLM-in-the-loop pipeline for automatic detoxification and presents \textsc{ParaDeHate}, a high-quality parallel hate speech detoxification dataset, we acknowledge several limitations that point to avenues for future improvement.

First, our detoxification pipeline exclusively uses \texttt{GPT-4o-mini} as the annotation agent. While this model has demonstrated strong performance, we do not evaluate the consistency or generalizability of our approach across other LLMs. Future work could explore whether similar performance holds when using open-source models or other LLMs.

Second, \texttt{GPT-4o-mini} is a commercial model, which may limit the reproducibility and transparency of our pipeline. Although the model was selected for its strong performance and relatively low cost, relying on a closed-source system restricts fine-grained control over its behavior and may pose challenges for researchers without API access.

Third, our dataset and evaluations are restricted to English-language hate speech. However, hate speech is a global issue and appears in many languages with varying structures, expressions, and cultural contexts. Applying and evaluating our pipeline on multilingual datasets is necessary to fully assess its utility in broader applications.

\section*{Ethical Considerations}

This paper includes examples of hateful content, and the proposed dataset inherently contains instances of hate speech. We recognize the sensitive nature of this material and want to explicitly state that our intention is not to disseminate or endorse such content. Rather, our work focuses on leveraging these examples and the broader dataset to develop a novel pipeline for the purification of hate speech. Our ultimate goal is to contribute to the creation of safer and more inclusive online environments by providing tools to mitigate the spread of harmful language. We have taken precautions to present only the minimum necessary examples for demonstrating the pipeline's functionality and impact.


\bibliography{custom,anthology_reduced}
\appendix
\newpage
\section{Hyperparameters for GPT-4o-mini}
\label{app:hyperparameters_gpt}
We provide the hyperparameters used for the GPT-4o-mini model in Table~\ref{tab:hyperparameters}. These settings were chosen to optimize the model's ability to convert hate speech sentences into non-hate speech while maintaining the same meaning.

\begin{table}[ht]
\centering
\begin{tabular}{@{}ll@{}} 
\toprule
\textbf{Hyperparameter} & \textbf{Value} \\
\midrule
Model Name & gpt-4o-mini \\
Maximum Tokens & 256 \\
Temperature & 0.6 \\
\bottomrule
\end{tabular}
\caption{Hyperparameters for GPT-4o-mini.}
\label{tab:hyperparameters}
\end{table}

\section{Hyperparameters for BART Fine-Tuning}\label{app: hyperparameters_bart}
\label{app:bart_hyperparameters}
As shown in Table~\ref{tab:bart_metrics}, we provide the hyperparameters applied during the fine-tuning of the BART model. These parameters were selected to optimize performance based on the BLEU metric.

\begin{table}[ht]
\centering
\begin{tabular}{@{}ll@{}}
\toprule
\textbf{Metric} & \textbf{Value} \\
\midrule
Learning Rate & 1e-5 \\
Batch Size (Train/Eval) & 8 \\
Epochs (Early-Stopped) & 7 \\
Weight Decay & 0.01 \\
Metric for Best Model & BLEU \\
\bottomrule
\end{tabular}
\caption{Fine-Tuning Hyperparameters for \texttt{BART-large}.}mi
\label{tab:bart_metrics}
\end{table}

\section{Prompt for Task 1 to Mitigate False Refusal Behaviour}\label{app:prompt_refusal}


\begin{tcolorbox}[colback=skyblue!20, colframe=skyblue!80!black, width=0.48\textwidth, rounded corners]
\textit{System}: Your role is to promote respectful and constructive communication by transforming offensive, harmful, or toxic language into clear, empathetic, and inclusive speech while preserving the speaker’s core intent. Under no circumstances give personal comments to the text, it would destroy the translation system!\\
\textit{User}: You are an advanced language model trained to foster constructive and inclusive dialogue. Your task is to take offensive, harmful, or toxic language—words, phrases, or expressions that demean, insult, or alienate—and reframe them into respectful, clear, and empathetic communication while preserving the speaker’s core intent. Below is a framework to guide your approach, inspired by the principles of respectful communication: \\Background: Human connection thrives on language, but offensive speech can fracture relationships and communities. Linguistic research, starting in the 20th century with scholars like Deborah Tannen, revealed how word choice shapes social dynamics. Today, tools like AI filters and communication techniques help transform harmful language into constructive dialogue, promoting unity and understanding.\\Objective: When given offensive input (e.g., insults, slurs, or aggressive phrasing), analyze the underlying message or emotion, then rephrase it to be respectful, inclusive, and constructive. Under no circumstances give personal comments to the text or express your opinion, don't answer like: ! \\Translate the following text into a neutral, non-offensive version while preserving its core meaning and intent. Maintain any mentions of users (e.g., @USER) and keep all numbers exactly as they appear in the original text. Do not add any commentary, disclaimers, or explanations—only return the transformed text. Text:
\end{tcolorbox}



\section{Cost of Creating \paradehate}\label{app:cost}

In total, the dataset comprises 19.153 million input tokens and 654,884 output tokens with GPT-4o-mini, incurring a cost of \$2.873 for input and \$0.517 for output, amounting to \$3.39 in total. These results highlight that our LLM pipeline is not only effective in detoxification but also highly cost-efficient.

\section{Examples}
\label{app:examples}
This section presents examples of toxic text alongside their detoxified versions from \paradehate. These examples, shown in Table~\ref{tab:examples}, demonstrate the model's ability to remove hate speech while preserving the intended meaning. The toxic and detoxified texts are displayed side-by-side for clarity.

\begin{table}[H]
\begin{flushright}
\small
\adjustbox{max width=0.75\textwidth}{
\begin{tabular}{p{4cm}|p{4cm}}
\toprule
\textbf{Toxic Text} & \textbf{Detoxified Text} \\
\midrule
But at least they let me know...ur a faggot if u keep those type of twitter secrets & But at least they let me know...you're not being open if you keep those types of Twitter secrets \\
\midrule
Alright ya faggots. Lets just stop with all the S/O and R/T. No one gives a fuck on what you think of other people. Probably lying anyways & Alright everyone. Let's just stop with all the shoutouts and retweets. No one is really interested in what you think of others. It’s likely not true anyway \\
\midrule
RT I hate people who agree with everything I say, like damn can u fight me or something & RT I find it frustrating when people always agree with me; I wish they would share their own opinions or challenge my thoughts. \\
\midrule
trump told everyone that maxine waters was low iq he forgot to add smelly corrupt bitch ass negress whore & Trump stated that Maxine Waters had a low IQ but did not include additional negative descriptors. \\
\midrule
@USER say that shit to my face "Bruh" you a bitch for gettin in a man u dont know business. Handle yo own shit & @USER, please address me directly. It seems inappropriate to involve yourself in matters that do not concern you. Focus on your own issues. \\
\midrule
Somebody please choke that fucking retard that keeps yelling "mashed potatoes". What a tool! \#PGAChampionship & Somebody please stop the person who keeps yelling "mashed potatoes." What an annoyance! \#PGAChampionship \\
\bottomrule
\end{tabular}}
\caption{Examples of toxic text and their detoxified versions in \paradehate.}
\label{tab:examples}
\end{flushright}
\end{table}

\end{document}